\documentclass[conference]{IEEEtran}
\IEEEoverridecommandlockouts
\usepackage[
    letterpaper,
    top=1.9cm,
    bottom=2.54cm,
    left=1.57cm,
    right=1.57cm
]{geometry}
\usepackage{cite}
\usepackage{amsmath,amssymb,amsfonts}
\usepackage{algorithmic}
\usepackage{graphicx}
\usepackage{textcomp}
\usepackage{xcolor}
\usepackage{soul}
\usepackage{booktabs}
\usepackage{float}

\usepackage{multirow}
\usepackage{url}
\usepackage{hyperref}
\hypersetup{
    colorlinks=true, 
    linkcolor=black, 
    citecolor=black, 
    urlcolor=black 
}
\def\BibTeX{{\rm B\kern-.05em{\sc i\kern-.025em b}\kern-.08em
    T\kern-.1667em\lower.7ex\hbox{E}\kern-.125emX}}

\begin{document}

\title{Evolutionary Physics-Informed Temporal Fusion for Lane-Change Intention Prediction}

\author{\IEEEauthorblockN{1\textsuperscript{st} Jiazhao Shi*}
\IEEEauthorblockA{\textit{Tandon School of Engineering} \\
\textit{New York University}\\
New York, USA \\
js12624@nyu.edu}
\and
\IEEEauthorblockN{2\textsuperscript{nd} Qiyang Xie}
\IEEEauthorblockA{\textit{Khoury College of Computer Science} \\
\textit{Northeastern University}\\
Boston, USA \\
xie.qiy@northeastern.edu}
\and
\IEEEauthorblockN{3\textsuperscript{rd} Ziyu Wang}
\IEEEauthorblockA{\textit{School of Business} \\
\textit{Wake Forest University}\\
Winston Salem, USA \\
martin\_ziyu@outlook.com}
\and
\IEEEauthorblockN{4\textsuperscript{th} Yichen Lin}
\IEEEauthorblockA{\textit{Tandon School of Engineering} \\
\textit{New York University}\\
New York, USA \\
ycl355@nyu.edu}
\and
\IEEEauthorblockN{5\textsuperscript{th} Di Zhu}
\IEEEauthorblockA{\textit{School of Engineering} \\
\textit{Santa Clara University}\\
Pittsburgh, USA \\
dizhu.judy.ml@gmail.com}
\and
\IEEEauthorblockN{6\textsuperscript{th} Chen Xie}
\IEEEauthorblockA{\textit{Manning College of Information and Computer Sciences} \\
\textit{University of Massachusetts Amherst}\\
Amherst, USA \\
chen678x@gmail.com}
\and
\IEEEauthorblockN{7\textsuperscript{th} Ziwei Wang}
\IEEEauthorblockA{\textit{College of Engineering} \\
\textit{Carnegie Mellon University}\\
Pittsburgh, USA \\
waltzvee@gmail.com}
\and
\IEEEauthorblockN{8\textsuperscript{th} Haoyun Zhang}
\IEEEauthorblockA{\textit{School of Arts and Sciences} \\
\textit{University of Pennsylvania}\\
Philadelphia, USA \\
haoyunz@sas.upenn.edu}
\and
\IEEEauthorblockN{9\textsuperscript{th} Enliang Li}
\IEEEauthorblockA{\textit{Qualcomm CDMA Technologies} \\
\textit{Qualcomm Technologies, Inc}\\
San Diego, USA \\
enlili@qti.qualcomm.com}
\and
\IEEEauthorblockN{10\textsuperscript{th} Zetong Guan}
\IEEEauthorblockA{\textit{College of Engineering} \\
\textit{University of Michigan}\\
Ann Arbor, USA \\
ztguan@umich.edu}
}

\maketitle

\begin{abstract}
Early lane-change intention prediction is essential for autonomous driving and ADAS, but it remains challenging because lane-changing behavior depends on evolving traffic risk, surrounding-vehicle interactions, and target-lane feasibility rather than only instantaneous vehicle states. This study proposes an evolutionary physics-informed temporal fusion framework for three-class lane-change intention prediction, including left lane change, right lane change, and no lane change. Instead of using static physics-informed variables alone, the proposed method derives temporal descriptors from conventional traffic signals, including risk evolution, gap persistence, counterfactual lane utility, interaction pressure gradient, maneuver feasibility, and intent consistency. These descriptors are fused with temporal embeddings learned from raw trajectory sequences through a sequence encoder, and the fused representation is used for final classification. Experiments are conducted on the highD and exiD datasets under 1\,s, 2\,s, and 3\,s prediction horizons. The proposed model achieves Macro F1-scores of 0.9514, 0.9256, and 0.8872 on highD, and 0.9386, 0.9070, and 0.8531 on exiD, respectively. The improvement is especially pronounced in exiD ramp-adjacent scenarios, indicating that temporal physical evolution is particularly useful in interaction-rich environments. These results demonstrate that combining evolutionary physics-informed descriptors with learned temporal representations provides a more dynamic and interpretable solution for early lane-change intention prediction.
\end{abstract}


\section{Introduction}

\subsection{Background and Motivation}

Lane-change intention prediction is a fundamental problem in autonomous driving systems and Advanced Driver Assistance Systems (ADAS), because lane-changing behavior directly affects motion planning, collision avoidance, and risk-aware decision-making. Unlike steady lane keeping, a lane-change maneuver is not determined solely by the ego vehicle's instantaneous motion state. Instead, it emerges from a dynamic decision process involving recent ego-vehicle motion, surrounding-vehicle interactions, adjacent-lane opportunities, and the driver's response to changing traffic constraints. Classical lane-changing theory has emphasized that lane-change decisions are governed by both incentive and safety considerations, including the expected benefit of moving to a target lane and the possible risk imposed on neighboring vehicles~\cite{zheng2014recent, kesting2007general}.

With the availability of high-resolution naturalistic trajectory datasets, learning-based lane-change prediction has become an active research direction. Existing studies have incorporated trajectory data, surrounding-vehicle context, and road-environment information to predict lane-change decisions or future trajectories~\cite{xue2022integrated}. The exiD dataset further extends highway trajectory research from regular straight-road segments to highly interactive highway entry and exit areas, providing a more challenging benchmark for modeling merging, diverging, and ramp-adjacent behaviors~\cite{moers2022exid}. These developments suggest that lane-change prediction should be treated not only as a motion-pattern recognition problem, but also as a context-dependent temporal reasoning problem.

Early lane-change prediction is especially challenging because useful predictions must be made before the vehicle crosses the lane boundary or fully enters the target lane. At longer prediction horizons, however, observable intention cues are often weak, incomplete, or ambiguous. Slight lateral drift, small heading variation, or gradual speed adjustment may indicate an emerging maneuver, but such cues can also be confused with normal lane-keeping fluctuations, especially in dense traffic. Therefore, early lane-change intention prediction requires models that capture not only the current traffic state, but also how maneuver-related physical signals evolve over time.

Our previous study proposed a physics-informed AI framework for three-class lane-change intention prediction across multiple highway scenarios, demonstrating that domain-informed variables can improve interpretability and predictive performance in both straight-road and ramp-related environments~\cite{shi2026multiscenario}. However, that work mainly used physics-informed variables as structured representations of the traffic state. Although such features are useful, they primarily describe what the physical state is at a given moment rather than how the state evolves before a lane-change decision becomes explicit. This motivates the present study, which reformulates physics-informed lane-change prediction as a temporal evolution modeling problem.

\subsection{Limitations of Static Physics-Informed Modeling}

Physics-informed variables such as Time-to-Collision (TTC), Time Headway (THW), Distance Headway (DHW), relative distance, relative velocity, lane offset, and target-lane gap size provide interpretable descriptions of motion dynamics, safety margins, and interaction feasibility. Recent lane-change risk and behavior-prediction studies continue to show the value of integrating traffic context, driving style, and interpretable risk-related variables into learning-based models~\cite{ren2025lanechange, huang2024driver}.

However, many conventional physics-informed variables are static or only weakly temporal when used directly. A single TTC value describes the estimated collision time at one moment, but it does not indicate whether risk is worsening, recovering, or fluctuating. Similarly, a target-lane gap size may indicate that an adjacent-lane opportunity exists at the current frame, but it does not reveal whether the gap is opening, closing, or likely to remain feasible long enough for a maneuver. Therefore, the key predictive evidence for early lane-change intention may lie not only in the instantaneous value of a physical variable, but also in its temporal behavior over the observation window.

This limitation is particularly important in complex traffic scenarios. In relatively regular straight-road environments, static traffic-state variables may already provide strong predictive cues. In ramp-adjacent environments, however, vehicles face changing lane topology, merging and diverging flows, and rapidly evolving interaction constraints. Recent context-aware and topology-aware lane-change studies have highlighted the importance of structured interaction modeling for complex traffic scenarios~\cite{huang2024driver}. More broadly, recent spatiotemporal traffic modeling work also suggests that traffic prediction benefits from architectures that explicitly capture temporal dynamics and interaction structure rather than relying only on static representations~\cite{li2026fast}.

\subsection{Core Idea: Evolutionary Physics-Informed Temporal Modeling}

This study reframes physics-informed lane-change prediction as a temporal evolution modeling problem rather than a static feature classification problem. Instead of using TTC, THW, DHW, gap size, relative motion, and lane offset only as instantaneous inputs, we derive evolutionary physics-informed descriptors that summarize how risk, opportunity, interaction pressure, and maneuver feasibility change within the observation window.

Specifically, the proposed descriptor system includes six groups: risk evolution, gap persistence, counterfactual lane utility, interaction pressure gradient, maneuver feasibility, and intent consistency. These descriptors represent whether safety risk is deteriorating or recovering, whether a target-lane opportunity is stable or transient, whether moving left or right would reduce projected risk, whether surrounding-vehicle pressure changes directionally, whether the maneuver is physically feasible, and whether multiple weak cues support the same maneuver direction. In this way, physically meaningful traffic signals are reinterpreted as dynamic evidence of emerging lane-change intention rather than as fixed snapshots of the traffic state.

To further exploit temporal information, this study develops a temporal-physics fusion framework for three-class lane-change intention prediction. The temporal branch learns latent maneuver representations from raw trajectory sequences, while the physics branch constructs evolutionary descriptors from physically meaningful traffic variables. The two information streams are fused and passed to a classifier to predict left lane change, right lane change, or no lane change.

This design combines the representation capacity of sequence learning with the interpretability and robustness of physics-informed modeling. Recent studies on lane-change maneuver and risk prediction have shown the value of jointly modeling maneuver probability, risk level, and time-related behavior~\cite{yang2025predicting}, while sequence models such as LSTM-based and Transformer-based architectures remain widely used for temporal driving-behavior prediction~\cite{fu2023lstm, lu2026knowlcp}. In this study, however, temporal learning is not used as a standalone replacement for physics-informed modeling. Instead, learned temporal embeddings are integrated with evolutionary physical descriptors to form a unified temporal-physics representation.

\subsection{Main Contributions}

The main contributions of this study are summarized as follows.

First, we propose an evolutionary physics-informed descriptor system for lane-change intention prediction. Different from static physics-informed features used in prior work, the proposed descriptors transform conventional traffic variables into temporal representations of risk evolution, gap persistence, counterfactual lane utility, interaction pressure gradient, maneuver feasibility, and intent consistency. This design enables the model to capture both the current traffic state and the direction and stability of its short-term evolution.

Second, we develop a temporal-physics fusion model that combines sequence-based temporal representation learning with evolutionary physics-informed descriptors. The temporal branch extracts latent maneuver dynamics from raw trajectory sequences, while the physics branch encodes interpretable evidence of evolving risk, opportunity, interaction pressure, and feasibility. To clarify the role of each component, the proposed model is compared with standalone LightGBM, standalone Bi-LSTM, standalone Transformer, and controlled feature-based variants.

Third, we evaluate the proposed framework on both highD and exiD under multiple prediction horizons. The proposed model achieves Macro F1-scores of 0.9514, 0.9256, and 0.8872 on highD and 0.9386, 0.9070, and 0.8531 on exiD for 1\,s, 2\,s, and 3\,s prediction horizons, respectively. The improvement is especially clear in the more complex exiD setting, where the proposed model improves Macro F1 by approximately 7.1 percentage points over the static LightGBM baseline at the 3\,s horizon. This supports the claim that evolutionary temporal physics descriptors are particularly beneficial in interaction-rich ramp, merging, and diverging scenarios~\cite{lin2017focal, cui2019classbalanced}.

Overall, this paper moves beyond the static physics-informed framework of our previous work~\cite{shi2026multiscenario} by making temporal evolution the central modeling target. By treating traffic-domain variables as evolving physical signals and integrating them with learned temporal representations, the proposed framework provides a more dynamic, interpretable, and scenario-aware solution for early lane-change intention prediction.

\section{Problem Formulation and Lane-Change Labeling}

\subsection{Three-Class Lane-Change Intention Prediction}

This study formulates lane-change intention prediction as a three-class classification problem, including left lane change (LLC), right lane change (RLC), and no lane change (NLC). For each ego vehicle, the model observes a historical trajectory segment within an observation window $W$ and predicts whether a lane-change maneuver will occur within a future prediction horizon $T$. Given the trajectory sequence $\mathbf{X}_{t-W:t}$, the prediction task is defined as:
\begin{equation}
\mathbf{X}_{t-W:t} \rightarrow y_{(t,t+T]}, \quad y_{(t,t+T]} \in \{\text{LLC}, \text{RLC}, \text{NLC}\}.
\end{equation}
Here, $W$ determines the amount of historical motion information available to the model, while $T$ determines how early the lane-change intention is predicted. The label indicates whether a left or right lane-change event occurs within the future horizon $(t,t+T]$; otherwise, the sample is labeled as NLC. A larger prediction horizon supports earlier warning but also increases uncertainty because lane-change cues are less explicit at earlier stages.

The definitions of lane-change events are specified according to the roadway structures of the highD and exiD datasets, which represent straight-road highway scenarios and ramp-adjacent merging/diverging scenarios, respectively. This distinction is necessary because lane identifiers are reliable for direction inference in highD but may not be sequential across mainline and ramp lanes in exiD. The detailed dataset characteristics are introduced later in Section~V.

\subsection{Straight-Road Scenarios in highD}

\begin{figure}[H]
\centering
\includegraphics[width=\columnwidth]{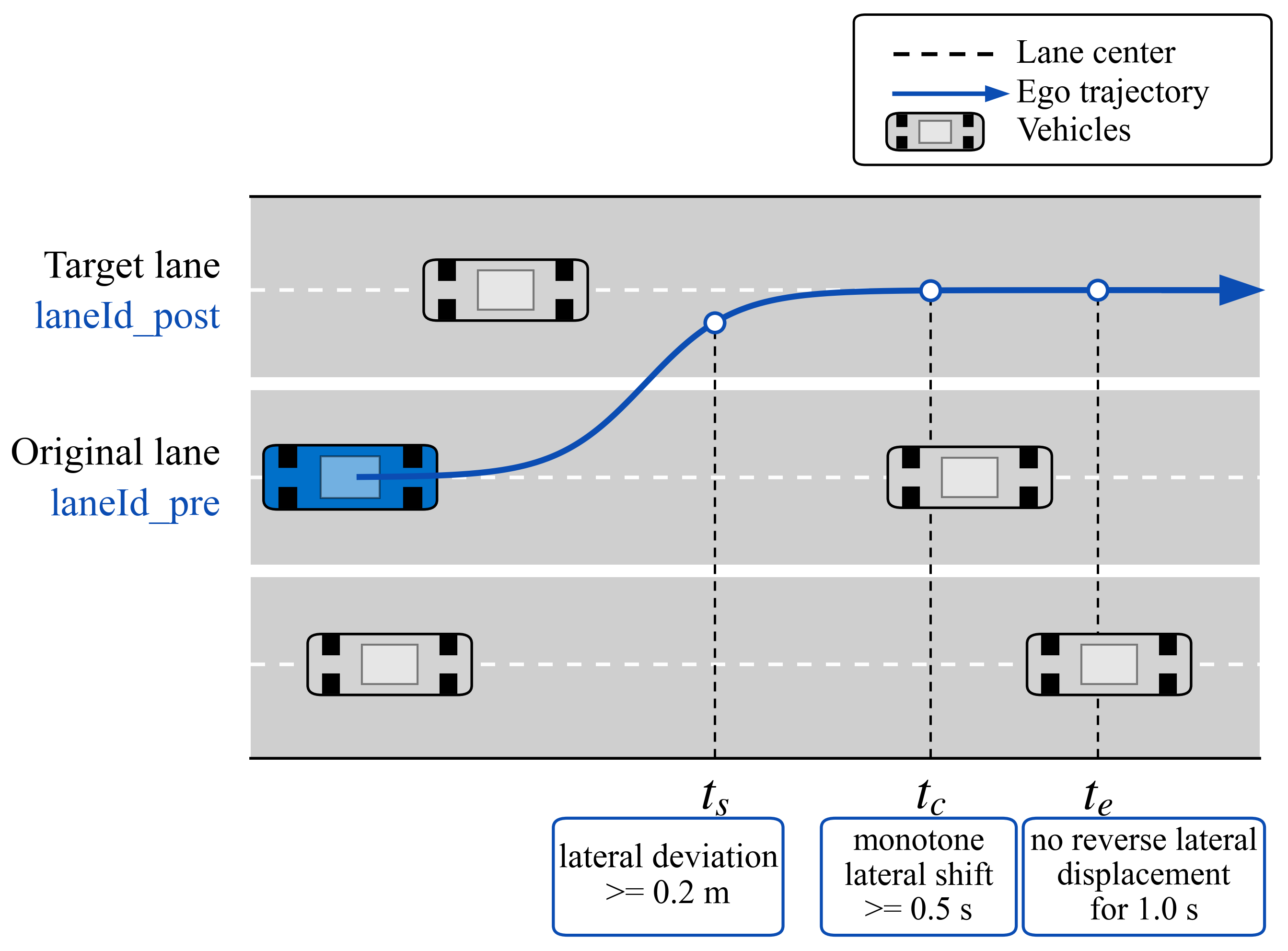}
\caption{Temporal labeling of straight-road lane-change events in highD.}
\label{fig:highd_labeling}
\end{figure}

Figure~\ref{fig:highd_labeling} illustrates the temporal labeling rule for a lane-change event on a straight roadway in the highD dataset.

Start time: The maneuver onset is identified when the vehicle deviates laterally by at least 0.2\,m from the lane center, with a continuous monotone lateral shift persisting for at least 0.5\,s thereafter.

End time: The maneuver is considered complete when the vehicle resides entirely within the target lane, with no reverse lateral displacement observed during the following 1.0\,s interval.

Direction: Since lane identifiers are sequential in highD straight-road scenarios, the maneuver direction is determined by comparing the laneId before and after the lane change. If the post-change laneId indicates movement toward the left side, the event is labeled as LLC; otherwise, it is labeled as RLC. Segments without a valid lane-change event during the prediction interval are labeled as NLC.

\subsection{Ramp Merging/Diverging Scenarios in exiD}

\begin{figure}[H]
\centering
\includegraphics[width=\columnwidth]{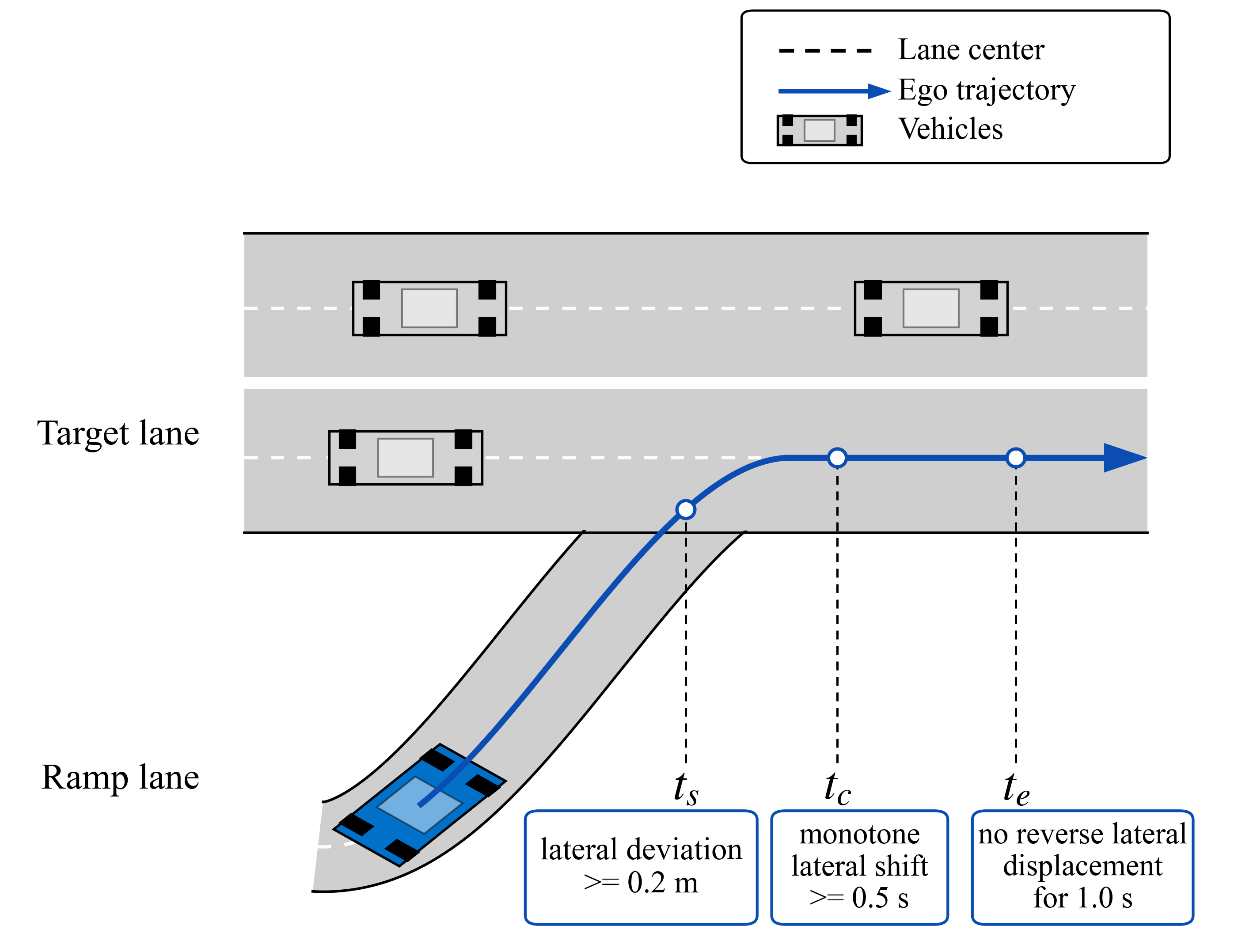}
\caption{Topology-aware labeling of ramp merging/diverging lane-change events in exiD.}
\label{fig:exid_labeling}
\end{figure}

Figure~\ref{fig:exid_labeling} illustrates the labeling rule for lane-change events in ramp merging and diverging areas in the exiD dataset.

Start time: The maneuver onset is identified using the same lateral-deviation criterion as highD: the vehicle must deviate laterally by at least 0.2\,m from the lane center, followed by a continuous monotone lateral shift for at least 0.5\,s.

End time: Completion is confirmed when the vehicle fully resides within the target lane and shows no reverse lateral displacement over the subsequent 1.0\,s interval.

Direction: Unlike highD, laneId values in exiD are not necessarily sequential between mainline lanes and ramp lanes. Therefore, lane-change direction is inferred from the vehicle's lateral velocity. Specifically, the mean lateral velocity is computed over a 0.1\,s interval beginning at the onset time. A positive mean lateral velocity indicates a left lane change, while a negative or non-positive value indicates a right lane change.

\section{Evolutionary Physics-Informed Descriptor Construction}

Based on the motivation introduced in Section~I-C, this section formalizes the construction of evolutionary physics-informed descriptors. For each ego vehicle, let $\mathbf{X}*{t-W:t}$ denote the historical observation sequence within the observation window $W$, where $t$ is the prediction reference time. The descriptor construction process extracts base physical trajectory variables from $\mathbf{X}*{t-W:t}$ and transforms them into higher-level temporal descriptors that summarize recent trend, stability, persistence, directional consistency, and feasibility implication.

The final descriptor vector is organized into six groups:
\begin{equation}
\begin{aligned}
\mathbf{p}_t =
[&\mathbf{p}_t^{\mathrm{risk}},
\mathbf{p}_t^{\mathrm{gap}},
\mathbf{p}_t^{\mathrm{utility}},
\mathbf{p}_t^{\mathrm{pressure}},\\
&\mathbf{p}_t^{\mathrm{feasibility}},
\mathbf{p}_t^{\mathrm{consistency}}].
\end{aligned}
\end{equation}
These groups correspond to risk evolution, gap persistence, counterfactual lane utility, interaction pressure gradient, maneuver feasibility, and intent-consistency / hesitation descriptors.

\subsection{Pre-Reference Definition and Leakage Prevention}

To make descriptor construction reproducible and prevent future-information leakage, all descriptors are defined with respect to the prediction reference time $t$. For each prediction sample, the model can only use the pre-reference observation window from $t-W$ to $t$, which contains trajectory states, lane-geometry information, and surrounding-vehicle states available before the prediction time. The future interval after $t$ is used only to assign the supervised lane-change label and is never used to construct the descriptor vector.

For any base signal, temporal statistics such as trend, volatility, duration, persistence, and consistency are computed only within the pre-reference observation window. When a descriptor requires a short-term projection, the projection is extrapolated from the current value and the trend estimated before $t$, rather than from the realized future trajectory. Therefore, maneuver feasibility and intent consistency are defined only by whether the pre-reference physical evidence supports a feasible and directionally consistent maneuver, not by whether a lane change actually occurs after $t$.

\subsection{Base Physical Trajectory Variables}

The base physical variables cover five types of trajectory information: ego-vehicle kinematics, lane-position dynamics, longitudinal safety indicators, surrounding-vehicle interaction states, and candidate-lane opportunity variables. Ego-vehicle kinematic variables include longitudinal and lateral velocity, longitudinal and lateral acceleration, scalar speed, heading angle, heading variation, yaw-rate-like heading change, and short-term displacement. Lane-position variables include lane offset, distance to the current lane center, distance to lane boundaries, normalized lateral position within the lane, and the rate of change of lane offset.

Longitudinal safety indicators include TTC, THW, and DHW. Surrounding-vehicle interaction variables include relative distance, relative speed, relative acceleration, relative longitudinal position, relative lateral position, and the availability of front, rear, left-front, left-rear, right-front, and right-rear vehicles. Candidate-lane opportunity variables describe the left and right adjacent lanes as possible maneuver targets, including target-lane front gap, target-lane rear gap, total available gap, gap balance, target-lane rear approach speed, target-lane front vehicle speed, and current-lane versus adjacent-lane spacing difference. These base variables are not presented as the main novelty; they serve as the physical signals from which evolutionary descriptors are derived.

\subsection{Risk Evolution Descriptors}

Risk evolution descriptors describe whether the ego vehicle's safety condition is improving, deteriorating, stable, or unstable within the observation window. For each prediction sample, let $t$ denote the prediction reference time, and let $[t-W,t]$ denote the historical observation window with length $W$. Within this window, $\tau$ denotes a discrete observation time step. For any physical signal $z(\tau)$, such as TTC, DHW, THW, speed, lane offset, or target-lane gap size, its temporal trend is estimated by fitting a linear slope over the observation window:
\begin{equation}
\text{Trend}(z) = \frac{\sum_{\tau=t-W}^{t}(\tau - \bar{\tau})(z(\tau) - \bar{z})}{\sum_{\tau=t-W}^{t}(\tau - \bar{\tau})^2}
\end{equation}
where $z(\tau)$ is the value of signal $z$ at time step $\tau$, $\bar{z}$ is the mean value of $z(\tau)$ within the observation window, and $\bar{\tau}$ is the mean of all time steps in the same window. The numerator measures the covariance between time and the signal value, while the denominator normalizes this covariance by the temporal variance. Therefore, $\text{Trend}(z)$ represents the dominant short-term changing direction of signal $z$: a positive value indicates an increasing trend, while a negative value indicates a decreasing trend.

In addition to trend, risk acceleration and risk volatility are constructed to describe whether risk deterioration is intensifying and whether the safety profile is stable or fluctuating. Risk volatility is computed from the variation of TTC, THW, DHW, and relative speed within the observation window. DHW deterioration duration, THW recovery rate, and minimum-risk timing are also included to distinguish persistent risk accumulation from temporary fluctuations.

Another descriptor is time-to-threshold, which estimates how soon a risk signal may cross a predefined boundary if its recent trend continues:
\begin{equation}
\text{TimeToThreshold}(z) = \frac{z(t) - \theta_z}{-\text{Trend}(z) + \epsilon}
\end{equation}
where $z(t)$ is the current value of signal $z$ at the prediction reference time $t$, $\theta_z$ is the predefined risk threshold for signal $z$, $\text{Trend}(z)$ is the estimated temporal slope of $z$ within $[t-W,t]$, and $\epsilon$ is a small positive constant used to avoid division by zero. The negative sign before $\text{Trend}(z)$ is used because many safety-related variables, such as TTC, THW, and DHW, become more critical when they decrease. A smaller positive value of $\text{TimeToThreshold}(z)$ indicates that the signal is expected to reach the risk threshold sooner.

\subsection{Gap Persistence Descriptors}

Gap persistence descriptors evaluate whether the candidate target-lane gap is stable enough to support a lane change. For each candidate target lane, the available gap is described using the front and rear spacing around the potential insertion position. Let $G_k(\tau)$ denote the available gap size in candidate lane $k$ at time step $\tau$, where $k$ represents the left or right candidate lane. The target gap opening rate and closing rate are derived from the temporal change of $G_k(\tau)$ within the observation window. A positive trend of $G_k(\tau)$ indicates that the target gap is expanding, while a negative trend indicates that the gap is shrinking.

Gap lifetime estimates how long the target-lane gap is expected to remain above a minimum feasible threshold. Let $G_k(t)$ denote the current gap size in candidate lane $k$, and let $\theta_G$ denote the minimum gap threshold required for a feasible lane change. If the gap is closing rapidly, the estimated lifetime becomes short; if the gap is stable or opening, the lifetime becomes longer. Gap persistence score summarizes whether the gap remains consistently feasible across the observation window or a short projection interval. Front-rear gap asymmetry and rear-approach threat are also included to describe whether the available space is balanced and whether the target-lane rear vehicle is approaching quickly.

\subsection{Counterfactual Lane-Utility Descriptors}

Counterfactual lane-utility descriptors compare the projected utility of staying in the current lane with the hypothetical utility of moving to the left or right lane. For each candidate action, namely staying in the current lane, changing to the left lane, or changing to the right lane, a projected risk score is computed using available physical signals. Let $k \in {\text{current}, \text{left}, \text{right}}$ denote the candidate lane or action being evaluated. A simplified projected risk score for lane $k$ is expressed as:
\begin{equation}
R_k = \alpha_1 R_k^{\text{front}} + \alpha_2 R_k^{\text{rear}} + \alpha_3 R_k^{\text{gap}} + \alpha_4 R_k^{\text{pressure}}
\end{equation}
where $R_k$ is the overall projected risk score associated with candidate lane $k$. The term $R_k^{\text{front}}$ represents the longitudinal risk caused by the front vehicle in lane $k$, $R_k^{\text{rear}}$ represents the risk caused by the rear vehicle in lane $k$, $R_k^{\text{gap}}$ represents the risk associated with the size and persistence of the available gap in lane $k$, and $R_k^{\text{pressure}}$ represents the surrounding interaction pressure around lane $k$. The coefficients $\alpha_1$, $\alpha_2$, $\alpha_3$, and $\alpha_4$ are non-negative weighting parameters controlling the relative contribution of the four risk components. A larger value of $R_k$ indicates a less favorable or more risky candidate lane.

Using this projected risk score, the left benefit gain is defined as the reduction in risk obtained by moving left instead of staying in the current lane, and the right benefit gain is defined similarly. If the projected risk of the left or right lane is lower than the current-lane risk, the corresponding benefit gain becomes positive; otherwise, the stay option remains more favorable. These descriptors connect physical risk evaluation with maneuver choice.

\subsection{Interaction Pressure Gradient Descriptors}

Interaction pressure gradient descriptors summarize how surrounding vehicles constrain the ego vehicle from different directions. Directional pressure is computed separately for the front, rear, left, and right regions around the ego vehicle. A general form of directional pressure is defined as:
\begin{equation}
P_q = \sum_{i \in \mathcal{N}_q} \frac{1 + \lambda |\Delta v_i|}{d_i + \epsilon}
\end{equation}
where $q \in {\text{front}, \text{rear}, \text{left}, \text{right}}$ denotes the spatial direction around the ego vehicle. $\mathcal{N}_q$ is the set of neighboring vehicles located in direction $q$, and $i$ indexes one neighboring vehicle in this set. The variable $d_i$ denotes the distance between the ego vehicle and neighboring vehicle $i$, while $\Delta v_i$ denotes their relative speed. Specifically, $|\Delta v_i|$ measures the magnitude of speed difference, so a larger value indicates stronger relative motion. The parameter $\lambda$ controls how much relative speed contributes to interaction pressure, and $\epsilon$ is a small positive constant used to avoid division by zero when the distance is very small. Therefore, $P_q$ becomes larger when nearby vehicles exist in direction $q$ or when those vehicles have strong relative motion with respect to the ego vehicle.

From these directional pressures, front pressure, rear pressure, left pressure, and right pressure are derived. The pressure change rate describes whether pressure from each direction is increasing or decreasing within the observation window. The left-right pressure gradient compares the pressure on the two sides of the ego vehicle, while front-rear pressure asymmetry compares the pressure ahead and behind the ego vehicle. These descriptors identify from which direction surrounding-vehicle constraints are increasing and whether either adjacent lane provides a lower-pressure alternative.

\subsection{Maneuver Feasibility Descriptors}

Maneuver feasibility descriptors evaluate whether the ego vehicle can physically complete a lane change under its current lateral motion and target-lane opportunity. The required lateral acceleration measures how much lateral acceleration would be needed for the ego vehicle to reach the target lane center within a reasonable completion time. Time-to-lane-center estimates how long the vehicle would need to reach the target lane center under its current lateral velocity, and time-to-boundary estimates how soon the vehicle will reach the lane boundary if the current lateral movement continues.

Time-to-complete-lane-change provides an estimate of how long the maneuver would take based on lateral distance, lateral velocity, and lateral acceleration. This estimate is compared with the target gap lifetime. The feasibility margin is defined as:
\begin{equation}
\text{FeasibilityMargin}_k = \text{GapLifetime}_k - \text{TimeToComplete}_k
\end{equation}
where $k$ denotes the candidate target lane, such as the left lane or the right lane. $\text{GapLifetime}_k$ is the estimated time duration during which the gap in lane $k$ is expected to remain feasible, and $\text{TimeToComplete}_k$ is the estimated time required for the ego vehicle to complete a lane change into lane $k$. Therefore, $\text{FeasibilityMargin}_k$ measures the time surplus between the available gap duration and the required maneuver completion time.

A positive feasibility margin indicates that the target gap is expected to remain available long enough for the vehicle to complete the lane change, while a negative margin indicates that the gap may disappear before completion. This descriptor directly connects vehicle motion capability with gap persistence.

\subsection{Intent-Consistency and Hesitation Descriptors}

Intent-consistency descriptors evaluate whether multiple weak cues support the same maneuver direction. Lateral-heading consistency measures whether lateral velocity and heading variation point in the same direction. Gap-intent consistency measures whether the vehicle is moving toward the side with a more persistent target gap. Risk-intent consistency measures whether the vehicle is moving toward the side with greater projected risk reduction. For each candidate direction, these descriptors summarize whether motion, gap, and utility cues support the same left or right maneuver tendency.

Hesitation descriptors capture unstable or aborted maneuver patterns, including lateral velocity sign changes, heading reversal, offset reversal, initial drift followed by return toward the lane center, and target gap loss after an initial opening. A high number of reversal cues indicates that the vehicle may not be committing to a lane change, even if short-term lateral movement is observed. These descriptors help distinguish true lane-change preparation from ordinary lane-keeping fluctuation or aborted maneuvers.

\subsection{Descriptor Normalization and Dataset-Specific Handling}

After construction, all descriptors are normalized using training-set statistics so that variables with different units and scales can be used jointly. Continuous descriptors, such as time-to-threshold, pressure, gap lifetime, feasibility margin, and lateral acceleration, are standardized using the training-set mean and standard deviation. Categorical or availability-related information, such as whether a surrounding vehicle exists in a specific region, is retained as binary indicators when needed.

Missing surrounding vehicles are handled according to their physical meaning. If no neighboring vehicle exists in a given direction, the corresponding pressure is set to zero because there is no immediate interaction pressure from that side. Distance-based variables are assigned a capped maximum value to avoid unrealistically large numbers. Availability indicators are used so that the model can distinguish between a genuinely low-pressure situation and a missing-neighbor case.

All rolling-window descriptors are computed only from $\mathbf{X}_{t-W:t}$. The future trajectory segment $(t,t+T]$ is used exclusively for assigning the supervised label and is never used for descriptor construction, parameter estimation, normalization, or threshold calibration. No validation or test trajectory information is used to tune these thresholds. In particular, maneuver feasibility is based on estimated completion time from the pre-reference lateral state, and intent consistency is based on agreement among pre-reference motion, gap, and utility cues rather than on the realized future lane-change outcome.

Some descriptor definitions are shared across highD and exiD, such as TTC evolution, gap persistence, pressure gradient, and feasibility margin. However, dataset-specific roadway structures are considered when computing lane geometry and candidate-lane availability. In highD, adjacent-lane candidates are defined mainly in straight highway segments. In exiD, ramp, merging, and diverging structures require candidate-lane availability to be checked with respect to local topology. This ensures that descriptors remain physically meaningful across both straight-road and ramp-related scenarios.

\section{Temporal-Physics Fusion Model}

This section presents the proposed temporal-physics fusion model for three-class lane-change intention prediction. The model is designed to combine two complementary sources of information: latent temporal representations learned from raw trajectory sequences and evolutionary physics-informed descriptors constructed from the temporal behavior of physical driving signals. The temporal encoder captures sequential maneuver patterns, while the descriptor branch provides interpretable physical summaries of evolving risk, gap opportunity, interaction pressure, and maneuver feasibility. The two representations are fused at the feature level and passed to a final classifier for prediction.

\subsection{Overall Architecture}
\begin{figure}[H]
\centering
\includegraphics[width=\columnwidth]{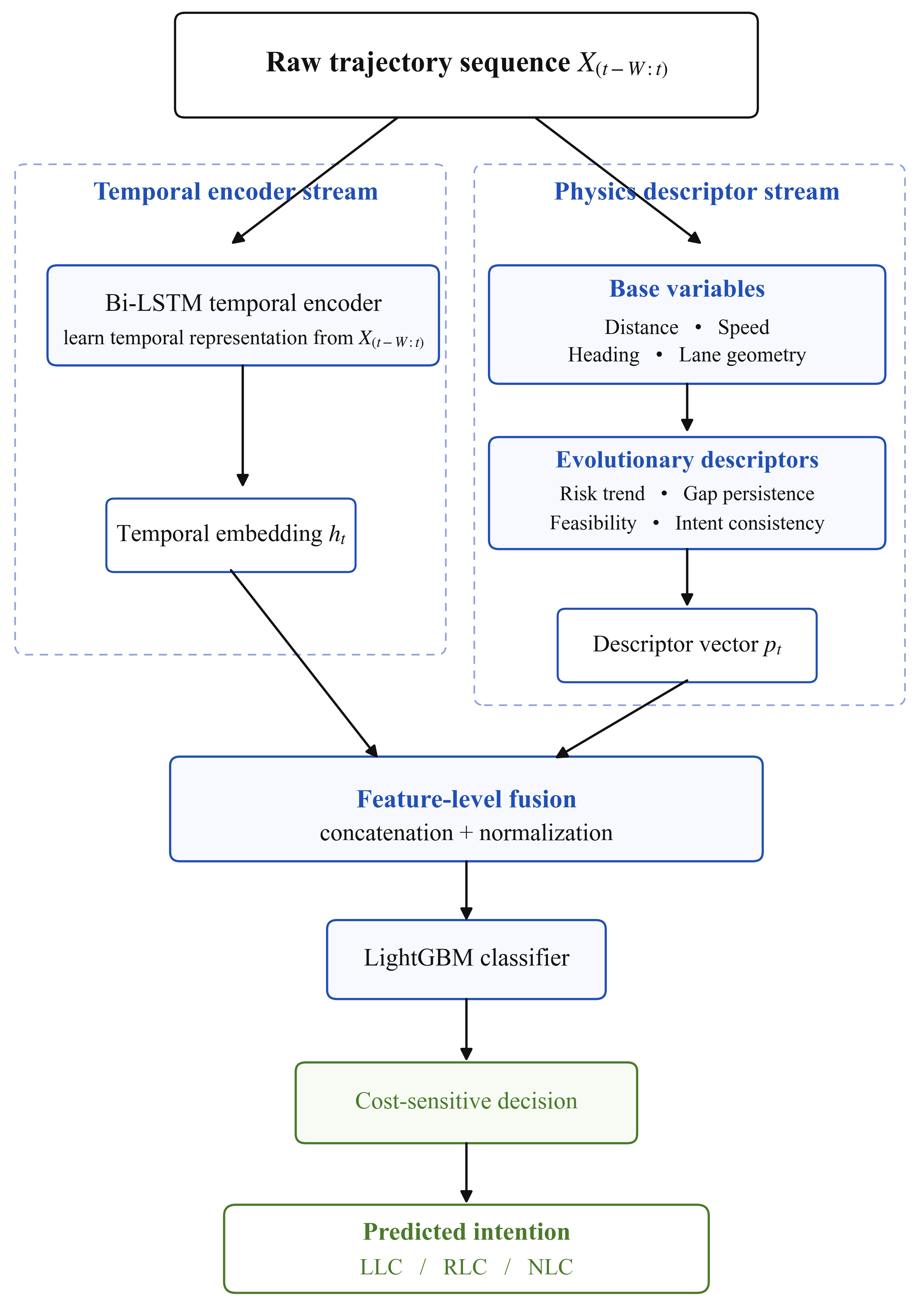}
\caption{Temporal-physics fusion architecture for lane-change intention prediction.}
\label{fig:architecture}
\end{figure}
Figure~\ref{fig:architecture} illustrates the overall architecture of the proposed temporal-physics fusion model. The input to the model is a historical trajectory sequence observed within the window $W$. This input is processed through two parallel streams. The first stream is the temporal encoder stream, which learns a compact temporal embedding from the raw trajectory sequence. The second stream is the evolutionary physics descriptor stream, which uses the descriptor construction procedure described in Section~III to generate a structured physics-informed vector.

The two streams are then fused into a unified representation. This fused vector is passed to a LightGBM classifier to predict the probability of each maneuver class. Finally, a cost-sensitive decision rule is applied to obtain the final predicted label among left lane change, right lane change, and no lane change.

The architecture can be summarized as follows:
\begin{align}
\mathbf{X}_{t-W:t} &\rightarrow \mathbf{h}_t, \\
\mathbf{X}_{t-W:t} &\rightarrow \mathbf{p}_t, \\
\mathbf{z}_t &= [\mathbf{h}_t \| \mathbf{p}_t], \\
\hat{y}_{t+T} &= f(\mathbf{z}_t),
\end{align}
where $\mathbf{X}_{t-W:t}$ is the observed trajectory sequence, $\mathbf{h}_t$ is the learned temporal embedding, $\mathbf{p}_t$ is the evolutionary physics-informed descriptor vector, $\|$ denotes feature concatenation, $\mathbf{z}_t$ is the fused representation, and $f(\cdot)$ is the final classifier.

This design differs from using a standalone temporal model or a standalone physics-feature model. The temporal encoder is not used as the final predictor by itself. Instead, it serves as a representation-learning module whose output is fused with physics-informed descriptors. In this way, the model can use both learned sequence patterns and interpretable physical evidence for prediction.

\subsection{Temporal Encoder}

The temporal encoder extracts latent maneuver representations from raw trajectory sequences. Given the input sequence $\mathbf{X}_{t-W}$, the encoder maps the ordered observations within the historical window into a fixed-length embedding vector $\mathbf{h}_t$. In this study, an LSTM or Bi-LSTM structure is used as the temporal encoder to capture sequential cues such as lateral drift, heading adjustment, acceleration variation, and interaction-induced motion changes.

For a Bi-LSTM encoder, the forward and backward hidden representations are combined to form the temporal embedding:
\begin{equation}
\mathbf{h}_t = [\overrightarrow{\mathbf{h}_t} | \overleftarrow{\mathbf{h}_t}]
\end{equation}
where $\overrightarrow{\mathbf{h}_t}$ and $\overleftarrow{\mathbf{h}_t}$ denote the final forward and backward hidden representations, respectively. The resulting embedding summarizes the temporal evolution of the observed trajectory segment. In the proposed framework, this encoder is not used as a standalone predictor; it provides a compact temporal representation that is fused with the evolutionary physics-informed descriptor vector before classification.

\subsection{Evolutionary Physics Descriptor Branch}

The evolutionary physics descriptor branch constructs a structured descriptor vector from the same historical trajectory segment $\mathbf{X}_{t-W:t}$. This branch uses the descriptor groups formalized in Section~III. Rather than directly feeding only instantaneous physical variables into the classifier, this branch summarizes the temporal behavior of physical signals within the observation window.

The output of this branch is denoted as:
\begin{equation}
\mathbf{p}_t \in \mathbb{R}^{d_p},
\end{equation}
where $d_p$ is the dimension of the evolutionary physics-informed descriptor vector. This vector provides interpretable information about the recent evolution of the traffic state and complements the latent temporal embedding learned by the sequence encoder.

\subsection{Feature-Level Fusion Strategy}

The proposed model uses feature-level fusion to combine the temporal embedding and the evolutionary physics-informed descriptor vector. Before fusion, both representations are normalized to improve numerical stability and reduce scale imbalance between learned neural embeddings and handcrafted physical descriptors.

Let $\mathbf{h}_t \in \mathbb{R}^{d_h}$ denote the temporal embedding generated by the temporal encoder, and let $\mathbf{p}_t \in \mathbb{R}^{d_p}$ denote the descriptor vector generated by the physics branch. The fused representation is obtained by concatenating the two vectors:
\begin{equation}
\mathbf{z}_t = [\mathbf{h}_t | \mathbf{p}_t],
\end{equation}
where:
\begin{equation}
\mathbf{z}_t \in \mathbb{R}^{d_h + d_p}.
\end{equation}

Here, $d_h$ is the embedding dimension selected through validation, and $d_p$ is the physics descriptor dimension determined by the number of constructed descriptors after preprocessing and dataset-specific alignment. The final fused vector $\mathbf{z}_t$ is used as the input to the downstream classifier.

\subsection{Final Classifier}

The final classifier is implemented using LightGBM~\cite{ke2017lightgbm}. LightGBM is suitable for this task because the fused representation contains heterogeneous information, including continuous physical descriptors, normalized temporal statistics, availability indicators, and neural sequence embeddings. Tree-based boosting can capture nonlinear interactions among these mixed-scale features, making it appropriate for the fused vector produced by the proposed framework.

Given the fused representation $\mathbf{z}_t$, the classifier outputs class probabilities:
\begin{equation}
\mathbf{P}_t = [P(\text{LLC}|\mathbf{z}_t),; P(\text{RLC}|\mathbf{z}_t),; P(\text{NLC}|\mathbf{z}_t)].
\end{equation}

The predicted class is then obtained from these probabilities using the decision strategy described in the next subsection.

\subsection{Imbalance-Aware Training and Decision Strategy}

Lane-change intention prediction is affected by class imbalance because no-lane-change samples are usually much more frequent than left- and right-lane-change samples. This imbalance can cause the model to favor the dominant NLC class and reduce recognition performance for the safety-critical LLC and RLC classes. In this study, imbalance handling is treated as an auxiliary component rather than the central contribution.

During temporal encoder training, balanced mini-batch sampling is used to increase the representation of minority lane-change classes. Instead of allowing each mini-batch to be dominated by NLC samples, samples from LLC, RLC, and NLC are drawn using controlled class proportions. This helps the temporal encoder learn more useful representations for rare lane-change behaviors. 

For the temporal branch, a class-balanced focal loss can be used to reduce the dominance of easy majority-class samples and emphasize difficult or underrepresented samples. The focal-loss mechanism assigns larger learning weight to hard examples, while class balancing compensates for differences in class frequency. This is especially useful when early lane-change cues are subtle and minority-class samples are sparse.

At the final prediction stage, a cost-sensitive decision rule is applied to the class probabilities output by LightGBM. Instead of always selecting the class with the highest probability, the decision rule can assign larger costs to missed LLC and RLC events. This reflects the safety-oriented nature of lane-change prediction, where missing an actual lane-change intention may be more serious than producing a conservative false alarm.

The final prediction is obtained by minimizing the expected classification cost:
\begin{equation}
\hat{y} = \arg\min_j \sum_i C_{ij}\, P(y=i|\mathbf{z}_t)
\end{equation}
where $C_{ij}$ denotes the cost of predicting class $j$ when the true class is $i$. Higher costs are assigned to missed lane-change events, encouraging the model to preserve recall for LLC and RLC.

Overall, the imbalance-aware strategy supports the temporal-physics fusion model by improving minority-class sensitivity. However, it is not the main methodological contribution. The primary contribution of the model lies in fusing learned temporal embeddings with evolutionary physics-informed descriptors for early lane-change intention prediction.

\section{Experimental Design}

This section describes the experimental design used to evaluate the proposed evolutionary physics-informed temporal fusion framework. The experiments are designed not only to measure predictive performance, but also to examine whether the proposed framework provides benefits beyond static physics-informed modeling, standalone sequence modeling, and standalone tree-based classification. Specifically, the evaluation focuses on three aspects: the contribution of evolutionary physics-informed descriptors, the contribution of temporal-physics fusion, and the robustness of the framework across roadway scenarios with different levels of interaction complexity.

\subsection{Datasets}

The proposed framework is evaluated on two large-scale naturalistic highway trajectory datasets: highD~\cite{krajewski2018highd} and exiD~\cite{moers2022exid}. These two datasets provide complementary testing environments for lane-change intention prediction.

The highD dataset contains drone-captured vehicle trajectories collected from straight highway segments. It provides high-resolution vehicle positions, velocities, accelerations, lane identifiers, and surrounding-vehicle information. Because highD mainly consists of regular highway segments without complex ramp topology, it is suitable for evaluating lane-change prediction under relatively structured straight-road conditions. In this setting, lane identifiers are generally sequential and can be used reliably to determine lane-change direction.

The exiD dataset extends the highD-style drone-based trajectory collection to highway entry, exit, merging, and diverging areas. Compared with highD, exiD contains more complex road topology and more interaction-intensive driving behaviors. Vehicles in exiD may interact with mainline traffic, ramp traffic, acceleration lanes, and deceleration lanes. Because lane identifiers may not be sequential across mainline and ramp lanes, lane-change direction in exiD requires topology-aware handling, as described in Section~II.

Using both datasets allows the experiments to evaluate the proposed framework under two different scenario types: relatively regular straight-road highway segments and more complex ramp-adjacent merging/diverging environments. This distinction is important because the proposed evolutionary descriptors and temporal fusion strategy are expected to be especially useful when traffic interactions and maneuver uncertainty are stronger.

\subsection{Data Preprocessing and Lane-Change Event Extraction}

Both highD and exiD were processed at the trajectory level before sample construction. For each vehicle, the raw trajectory records were first ordered by frame index, and vehicles with incomplete trajectory segments, missing lane information, or insufficient history length for the required observation window were removed. Basic kinematic variables, including longitudinal and lateral position, velocity, acceleration, and surrounding-vehicle information, were then extracted from the cleaned trajectories. To ensure temporal consistency, all features used for prediction were computed only from the historical observation segment before the prediction reference time.

Lane-change events were extracted according to the dataset-specific labeling rules defined in Section~II. For highD, lane-change direction was determined by the relative change in sequential lane identifiers. For exiD, because lane identifiers are not necessarily sequential across mainline and ramp lanes, lane-change direction was inferred using the topology-aware rule based on the lateral motion around the maneuver onset. Samples without any valid lane-change event within the prediction horizon were labeled as lane keeping.

To avoid ambiguous supervision, samples containing multiple lane-change events within the prediction horizon were discarded. This filtering ensures that each prediction sample corresponds to at most one future maneuver, so that the target label can be uniquely defined as no lane change, left lane change, or right lane change.

\subsection{Prediction-Window Construction and Class-Balance Treatment}

For each prediction sample, let $t$ denote the prediction reference time. The model observes only the historical trajectory segment within the observation window $[t-W,t]$, where $W$ is the observation length. The target label is determined from the future prediction horizon $(t,t+T]$. If a left or right lane-change event occurs within this future horizon, the sample is labeled according to the corresponding maneuver direction. If no lane-change event occurs within the horizon, the sample is labeled as lane keeping. Therefore, no future trajectory information beyond the label-generation horizon is used as model input.

Due to the sparse occurrence of lane-change maneuvers, the highD dataset exhibits severe class imbalance between lane-keeping and lane-changing samples. To mitigate this issue, we adopted a three-stage imbalance-handling strategy for highD. First, balanced mini-batch sampling was used during training to ensure that minority left- and right-lane-change samples were consistently represented in each batch. Second, a class-balanced focal loss was employed to increase the contribution of rare and difficult lane-change samples while down-weighting easy majority-class samples. Third, a cost-sensitive decision rule was applied during inference to assign higher penalties to missed lane-change detections than to false alarms. This strategy improves minority-class learning without generating synthetic trajectory samples. 

For exiD, no additional re-sampling was applied because the class distribution is more balanced. The original sample distribution was preserved, while the same model-selection and decision framework was used for fair comparison.

\subsection{Train/Validation/Test Split}
\label{sec:split}
To maintain a strict separation between training and evaluation data and to eliminate potential overlap in vehicle trajectories, both datasets were split according to recording locations rather than by randomly splitting individual samples. This location-level split prevents the same traffic scene or vehicle trajectory from appearing in multiple data partitions.

For highD, samples from locations 0, 1, 2, and 3 were assigned to the training pool, while samples from locations 4 and 5 were held out for final testing. For exiD, samples from locations 0, 1, 2, and 3 were used as the training pool, while samples from locations 4, 5, and 6 were used as the held-out test set. From the training pool, 20\% of the samples were further selected as the validation set, and the remaining 80\% were used for model training. The validation set was used only for hyperparameter tuning and model selection, while all final results were reported on the held-out test locations.

Because the split was performed at the recording-location level, no vehicle trajectory appears in more than one partition. This protocol enforces both vehicle-level and scenario-level separation between training, validation, and test data, thereby reducing the risk of trajectory leakage and overly optimistic evaluation results.

\subsection{Baseline Models and Model Variants}

To evaluate the effectiveness of the proposed temporal-physics fusion framework, we compare it with several baseline models and controlled variants. These models are designed to isolate the contributions of static physical variables, evolutionary physics-informed descriptors, sequence-based temporal learning, attention-based temporal learning, and feature-level fusion.

First, we use LightGBM with static physics features as a conventional physics-informed baseline. The input features include kinematic states, lane-position variables, TTC, THW, DHW, relative motion, and gap-related variables. This baseline evaluates whether static traffic-state features alone are sufficient for lane-change intention prediction.

Second, we use LightGBM with the proposed evolutionary physics-informed descriptors. This model removes the neural temporal encoder and relies only on the descriptor system introduced in Section~III. Comparing it with the static-feature LightGBM baseline shows whether the proposed descriptors provide additional predictive value beyond conventional static variables.

Third, we include standalone Bi-LSTM model using raw trajectory sequences. This model represent recurrent sequence-learning baselines without explicit evolutionary physics-informed descriptors. They are used to evaluate whether learned temporal representations alone can capture lane-change intention cues.

Fourth, we include a standalone Transformer model using raw trajectory sequences. This model follows an end-to-end sequence classification structure, where the historical trajectory sequence is projected into a latent embedding space, augmented with positional encoding, processed by stacked multi-head self-attention layers, and then passed to a classification head for lane-change intention prediction. The standalone Transformer does not use the proposed evolutionary physics-informed descriptors or the feature-level fusion module. Therefore, it serves as an independent attention-based temporal baseline rather than a controlled variant of the proposed framework.

Fifth, we evaluate a temporal encoder variant without the evolutionary descriptor branch. This model uses the same sequence encoder as the proposed framework but removes the physics-informed descriptor vector. It allows us to isolate whether the learned temporal embedding alone is sufficient for lane-change intention prediction.

Finally, the proposed temporal-physics fusion model combines the learned temporal embedding with the evolutionary physics-informed descriptor vector for final classification. To further analyze descriptor-level contributions, we conduct ablation experiments by removing representative descriptor groups, including counterfactual lane-utility descriptors and maneuver feasibility descriptors. Additional ablations on risk evolution, gap persistence, interaction pressure gradient, and intent-consistency descriptors are also included when space allows.

As summarized in Table~\ref{tab:baselines}, this baseline design supports a controlled evaluation of the proposed framework. The comparison between B1 and B2 evaluates whether evolutionary physics-informed descriptors provide additional value beyond static physics features. The comparison among B3, B4, and B5 examines whether recurrent, attention-based, and framework-specific temporal representations are sufficient without explicit evolutionary descriptors. The comparison between B5 and B6 isolates the contribution of feature-level fusion, while the comparison between B6 and the descriptor-ablation variants evaluates the contribution of individual descriptor groups.
\begin{table}[H]
\caption{Baseline Models and Controlled Variants.}
\label{tab:baselines}
\centering
\footnotesize
\begin{tabular}{@{}clp{3.8cm}@{}}
\toprule
\textbf{ID} & \textbf{Model} & \textbf{Purpose} \\
\midrule
B1 & LightGBM-static & Tests conventional static physics-informed modeling \\
B2 & LightGBM-evolutionary & Tests evolutionary descriptors without neural embeddings \\
B3 & Standalone Bi-LSTM & Tests recurrent sequence-based temporal learning \\
B4 & Standalone Transformer & Tests attention-based sequence classification without evolutionary descriptors \\
B5 & Temporal w/o descriptors & Tests the temporal branch alone \\
B6 & Proposed fusion & Tests the full framework \\
B7 & Descriptor-group ablations & Tests the contribution of each descriptor group\\
\bottomrule
\end{tabular}
\end{table}

\subsection{Evaluation Settings}

The dataset split follows the recording-location-level protocol described in Section~\ref{sec:split}. Validation performance is used to select the observation window length and tune model hyperparameters, while all final results are reported on the held-out test locations. No vehicle trajectory appears in more than one split, ensuring both vehicle-level separation and scenario-level separation between training and testing.

Model performance is evaluated under prediction horizons of 1\,s, 2\,s, and 3\,s before the lane-change onset. For each horizon, the model predicts one of three classes: left lane change, right lane change, or no lane change. Since the class distribution is imbalanced, especially because no-lane-change samples are more frequent, both overall accuracy and macro-averaged F1 score are reported. Accuracy measures the overall classification correctness, while macro-F1 gives equal importance to each class and better reflects the model's performance on minority lane-change classes.

All compared models are trained and evaluated using the same sample construction, data split, prediction horizons, and class definitions. Hyperparameters are selected based only on the validation set, and the held-out test set is used once for final reporting. This setting ensures a fair comparison between the proposed temporal-physics fusion model, standalone recurrent and Transformer-based temporal models, standalone physics-descriptor models, and baseline classifiers.

\subsection{Evaluation Objectives}

The experimental design aims to assess the proposed framework from four complementary perspectives. First, we evaluate whether the evolutionary physics-informed descriptors improve prediction performance compared with conventional static physics features. This comparison verifies whether modeling the temporal evolution of physical variables provides additional information beyond their instantaneous values.

Second, we examine whether the proposed temporal-physics fusion model outperforms standalone modeling strategies. This is done by comparing the full model with standalone LightGBM models, standalone Bi-LSTM models, and a standalone Transformer model. The purpose is to determine whether the fusion of learned temporal embeddings and interpretable physical descriptors provides a stronger representation than either physics-only modeling or end-to-end temporal sequence learning alone.

Third, we compare performance across highD and exiD to analyze scenario-dependent behavior. Since highD mainly represents straight-road highway segments while exiD contains ramp-adjacent merging and diverging scenarios, this comparison tests whether temporal-physics fusion provides larger benefits in more interaction-rich and structurally complex environments.

Finally, ablation experiments are used to examine the contribution of selected descriptor groups, especially counterfactual lane-utility descriptors and maneuver feasibility descriptors. These experiments help identify which parts of the evolutionary descriptor system are most useful for recognizing the minority LLC and RLC classes.

\section{Results}

This section presents the experimental results of the proposed evolutionary physics-informed temporal fusion framework. The analysis is organized to evaluate three major aspects of the proposed method. First, we compare the overall performance of the proposed model against standalone LightGBM, standalone Bi-LSTM, and standalone Transformer baselines. Second, we examine whether evolutionary physics-informed descriptors improve over static physics-informed features. Third, we analyze the effects of temporal fusion, dataset complexity, prediction horizon, and class-wise performance.

\subsection{Overall Comparison Across Models}

\begin{table}[H]
\caption{Overall Accuracy Performance Comparison.}
\label{tab:accuracy}
\centering
\scriptsize
\setlength{\tabcolsep}{3pt}
\begin{tabular}{@{}llccccc@{}}
\toprule
\textbf{Dataset} & \textbf{Horizon} & \textbf{LG-static} & \textbf{Bi-LSTM} & \textbf{Transformer} & \textbf{LG-evol.} & \textbf{Proposed} \\
\midrule
highD & 1\,s & 0.9981 & 0.9970 & 0.9977 & 0.9983 & \textbf{0.9986} \\
highD & 2\,s & 0.9952 & 0.9934 & 0.9928 & 0.9958 & \textbf{0.9962} \\
highD & 3\,s & 0.9906 & 0.9875 & 0.9859 & 0.9912 & \textbf{0.9918} \\
exiD  & 1\,s & 0.9607 & 0.8704 & 0.9318 & 0.9621 & \textbf{0.9637} \\
exiD  & 2\,s & 0.8950 & 0.7742 & 0.7626 & 0.9196 & \textbf{0.9386} \\
exiD  & 3\,s & 0.8117 & 0.7058 & 0.6904 & 0.8462 & \textbf{0.8759} \\
\bottomrule
\end{tabular}
\end{table}

\begin{table}[H]
\caption{Overall Macro F1 Performance Comparison.}
\label{tab:macrof1}
\centering
\scriptsize
\setlength{\tabcolsep}{3pt}
\begin{tabular}{@{}llccccc@{}}
\toprule
\textbf{Dataset} & \textbf{Horizon} & \textbf{LG-static} & \textbf{Bi-LSTM} & \textbf{Transformer} & \textbf{LG-evol.} & \textbf{Proposed} \\
\midrule
highD & 1\,s & 0.9359 & 0.9162 & 0.9287 & 0.9425 & \textbf{0.9514} \\
highD & 2\,s & 0.9126 & 0.8847 & 0.8829 & 0.9189 & \textbf{0.9256} \\
highD & 3\,s & 0.8761 & 0.8430 & 0.8264 & 0.8826 & \textbf{0.8872} \\
exiD  & 1\,s & 0.9004 & 0.7702 & 0.8526 & 0.9197 & \textbf{0.9386} \\
exiD  & 2\,s & 0.8343 & 0.7233 & 0.7148 & 0.8789 & \textbf{0.9070} \\
exiD  & 3\,s & 0.7819 & 0.6875 & 0.6521 & 0.8205 & \textbf{0.8531} \\
\bottomrule
\end{tabular}
\end{table}

Tables~\ref{tab:accuracy} and~\ref{tab:macrof1} summarize the overall performance of the main model families across highD and exiD under prediction horizons of 1\,s, 2\,s, and 3\,s. The compared models include LightGBM with static physics features, standalone Bi-LSTM using raw trajectory sequences, standalone Transformer using raw trajectory sequences, LightGBM with evolutionary physics-informed descriptors, and the proposed temporal-physics fusion model.

Overall, the proposed temporal-physics fusion model achieves the strongest performance across all datasets and prediction horizons. On highD, all models obtain high overall accuracy because the dataset mainly contains regular straight-road highway scenarios and the NLC class is highly dominant. However, Macro F1 reveals clearer differences among models. The proposed model improves Macro F1 over LightGBM-static from 0.9359 to 0.9514 at the 1\,s horizon, from 0.9126 to 0.9256 at the 2\,s horizon, and from 0.8761 to 0.8872 at the 3\,s horizon.

The standalone Transformer performs competitively at the 1\,s horizon, improving Macro F1 over Bi-LSTM from 0.9162 to 0.9287 on highD and from 0.7702 to 0.8526 on exiD. This suggests that self-attention can capture short-term temporal cues when the lane-change evidence is close to the maneuver onset. However, its performance decreases more noticeably at the 2\,s and 3\,s horizons, where early intention cues are weaker and less consistently reflected in raw trajectory sequences. In these longer-horizon settings, the Transformer falls below the Bi-LSTM model, indicating that a stronger attention-based architecture alone does not guarantee better early lane-change intention prediction.

The Transformer also remains clearly below LightGBM with evolutionary physics-informed descriptors, especially at longer horizons. This is reasonable because the Transformer learns temporal patterns directly from raw trajectory sequences, whereas LightGBM-evolutionary uses explicitly constructed descriptors that summarize risk evolution, gap persistence, interaction pressure, maneuver feasibility, and intent consistency. The comparison suggests that physically meaningful temporal descriptors can be more reliable than end-to-end sequence learning alone under imbalanced early-intention prediction settings.

The improvement of the proposed model is more pronounced on exiD. At the 3\,s prediction horizon, LightGBM-static achieves a Macro F1 of 0.7819, the standalone Transformer reaches 0.6521, and LightGBM-evolutionary reaches 0.8205, while the proposed fusion model further improves Macro F1 to 0.8531. This suggests that temporal-physics fusion provides greater benefits in ramp-adjacent scenarios where maneuver evolution and surrounding-vehicle interactions are more complex.

The standalone Bi-LSTM and Transformer baselines both perform worse than the proposed fusion model, especially on exiD and at longer prediction horizons. This does not mean that temporal learning is ineffective; rather, it shows that pure sequence learning without explicit physical descriptors may struggle when the training data are imbalanced and the maneuver evidence depends on interpretable physical interaction variables. The proposed model addresses this limitation by combining learned temporal embeddings with evolutionary physics-informed descriptors.

\subsection{Effect of Evolutionary Physics-Informed Descriptors}

To evaluate whether the proposed descriptors provide benefits beyond conventional static physics-informed features, we compare LightGBM-static and LightGBM-evolutionary. Both models use the same classifier family, so the difference mainly reflects the effect of replacing static feature inputs with evolutionary physics-informed descriptors. Table~\ref{tab:evo_effect} summarizes the Macro F1 comparison between static physics features and evolutionary physics-informed descriptors across different datasets and prediction horizons.

\begin{table}[H]
\caption{Effect of Evolutionary Descriptors vs.\ Static Features.}
\label{tab:evo_effect}
\centering
\footnotesize
\begin{tabular}{@{}llccc@{}}
\toprule
\textbf{Dataset} & \textbf{Horizon} & \textbf{Static F1} & \textbf{Evol.\ F1} & \textbf{$\Delta$} \\
\midrule
highD & 1\,s & 0.9359 & 0.9425 & +0.0066 \\
highD & 2\,s & 0.9126 & 0.9189 & +0.0063 \\
highD & 3\,s & 0.8761 & 0.8826 & +0.0065 \\
exiD  & 1\,s & 0.9004 & 0.9197 & +0.0193 \\
exiD  & 2\,s & 0.8343 & 0.8789 & +0.0446 \\
exiD  & 3\,s & 0.7819 & 0.8205 & +0.0386 \\
\bottomrule
\end{tabular}
\end{table}

The evolutionary descriptors consistently improve Macro F1 over static physics features on both datasets. The improvement is modest on highD, ranging from approximately 0.006 to 0.007 Macro F1. This is expected because straight-road lane changes in highD are more regular and can already be captured effectively by conventional kinematic, safety, and gap-related variables.

In contrast, the improvement is much larger on exiD. At the 2\,s horizon, the Macro F1 increases from 0.8343 to 0.8789. At the 3\,s horizon, it increases from 0.7819 to 0.8205. These gains indicate that modeling the temporal evolution of physical variables is especially useful in ramp merging and diverging areas. In such scenarios, instantaneous gap size or TTC values are often insufficient because the key predictive signal lies in whether risk is worsening, whether the target gap is persistent, and whether the surrounding-vehicle pressure is changing over time.

This result also explains why the LightGBM-evolutionary baseline outperforms the standalone Transformer in most settings. Although the Transformer provides an attention-based sequence-learning architecture, it must infer physical interaction patterns directly from raw trajectory sequences. This becomes more difficult at longer prediction horizons, where early maneuver cues are weaker and less stable. In contrast, the evolutionary descriptor representation directly encodes physically meaningful temporal summaries before classification. The result therefore supports the central claim of the proposed descriptor system: its main value does not come from simply adding more physical variables, but from transforming conventional physical signals into temporal descriptors that summarize their short-term evolution.

\subsection{Dataset-Specific Behavior: highD vs.\ exiD}

\begin{figure}[H]
\centering
\includegraphics[width=\columnwidth]{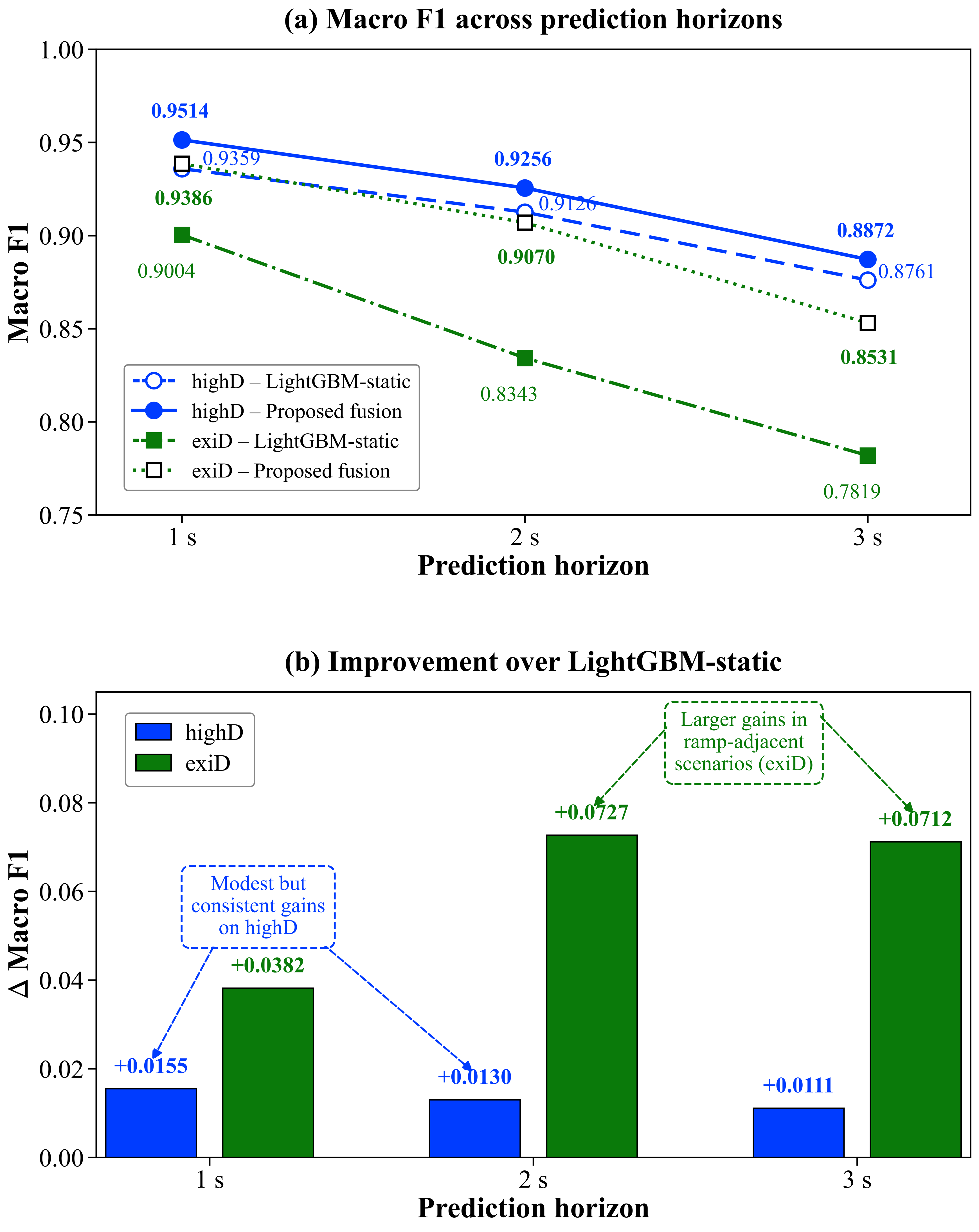}
\caption{Dataset-specific performance comparison between highD and exiD.}
\label{fig:dataset_comparison}
\end{figure}

The results reveal clear dataset-specific behavior. As shown in Figure~\ref{fig:dataset_comparison}, the proposed method improves performance over baseline models on highD, but the gain is relatively modest. This is mainly because highD contains straight-road highway scenarios where lane geometry is stable and surrounding-vehicle interactions are more regular. In such cases, static physical features and tree-based models already perform strongly, leaving less room for temporal fusion to provide large additional improvements.

On exiD, the gains are substantially larger. The exiD dataset contains ramp-adjacent merging and diverging scenarios, where vehicles must respond to changing lane topology, unstable gaps, and stronger surrounding-vehicle interactions. In these conditions, the temporal evolution of physical variables becomes more informative than their instantaneous values. The proposed fusion model is therefore more effective because it captures both explicit physical evolution and latent temporal sequence patterns.

\subsection{Prediction Horizon Analysis}

Across all model families and both datasets, performance decreases as the prediction horizon increases from 1\,s to 3\,s. This trend is consistent with the early prediction nature of the task. A longer horizon requires the model to predict the maneuver further in advance, when lane-change cues are weaker and less explicit.

On highD, the proposed fusion model maintains strong performance across all horizons. Its Macro F1 decreases from 0.9514 at 1\,s to 0.9256 at 2\,s and 0.8872 at 3\,s. Although the decline is visible, the model remains relatively stable because straight-road scenarios provide regular lane geometry and more predictable motion patterns.

On exiD, the decline is more pronounced. The proposed fusion model achieves a Macro F1 of 0.9386 at 1\,s, 0.9070 at 2\,s, and 0.8531 at 3\,s. This reflects the greater uncertainty of ramp merging and diverging behaviors. In such scenarios, a vehicle may initially show weak lane-change evidence but later adjust its plan due to changing gaps, surrounding-vehicle pressure, or ramp topology. Therefore, longer-horizon prediction is more difficult on exiD than on highD.

This horizon-based pattern also explains why standalone sequence models, especially the Transformer, degrade more clearly at longer horizons. When early cues are weak, raw trajectory sequences alone may not provide stable evidence for attention-based temporal learning. The proposed evolutionary descriptors help reduce this difficulty by explicitly capturing how risk, gap opportunity, and maneuver feasibility evolve before the actual maneuver.

\subsection{Ablation by Descriptor Group}

To evaluate the contribution of different evolutionary physics-informed descriptor groups, we conduct ablation experiments by removing one descriptor group at a time from the full temporal-physics fusion model. The temporal encoder and the remaining descriptor groups are kept unchanged. This setting allows us to examine how each descriptor group affects final prediction performance.

The ablation study focuses on Macro F1 and the F1-scores of the two minority lane-change classes, LLC and RLC. These metrics are more informative than overall accuracy because the NLC class is dominant and relatively easier to predict. The analysis is conducted under the 3\,s prediction horizon, where early prediction is most difficult and the value of temporal descriptors is expected to be more visible. Table~\ref{tab:ablation} summarizes the ablation results by descriptor group on highD and exiD.

\begin{table}[H]
\caption{Ablation Results at 3\,s Prediction Horizon.}
\label{tab:ablation}
\centering
\footnotesize
\begin{tabular}{@{}llccc@{}}
\toprule
\textbf{Dataset} & \textbf{Variant} & \textbf{Macro F1} & \textbf{LLC F1} & \textbf{RLC F1} \\
\midrule
highD & Full model & \textbf{0.8872} & \textbf{0.8252} & \textbf{0.8404} \\
highD & w/o risk evolution & 0.8786 & 0.8115 & 0.8281 \\
highD & w/o gap persistence & 0.8810 & 0.8172 & 0.8306 \\
highD & w/o counterfactual & 0.8794 & 0.8138 & 0.8279 \\
highD & w/o pressure gradient & 0.8823 & 0.8191 & 0.8328 \\
highD & w/o feasibility & 0.8801 & 0.8150 & 0.8312 \\
highD & w/o intent consistency & 0.8835 & 0.8204 & 0.8346 \\
\midrule
exiD & Full model & \textbf{0.8531} & \textbf{0.7924} & \textbf{0.8152} \\
exiD & w/o risk evolution & 0.8368 & 0.7706 & 0.7931 \\
exiD & w/o gap persistence & 0.8315 & 0.7628 & 0.7874 \\
exiD & w/o counterfactual & 0.8279 & 0.7552 & 0.7826 \\
exiD & w/o pressure gradient & 0.8346 & 0.7687 & 0.7895 \\
exiD & w/o feasibility & 0.8258 & 0.7521 & 0.7793 \\
exiD & w/o intent consistency & 0.8402 & 0.7765 & 0.7980 \\
\bottomrule
\end{tabular}
\end{table}

The ablation results show that removing any descriptor group reduces performance, indicating that the descriptor groups provide complementary information. The drop is relatively modest on highD, where lane-change behavior is more regular. In contrast, the degradation is larger on exiD, especially when removing maneuver feasibility, counterfactual lane utility, and gap persistence descriptors.

Among the descriptor groups, maneuver feasibility and counterfactual utility show the strongest effects on exiD. Removing feasibility descriptors reduces Macro F1 from 0.8531 to 0.8258, while removing counterfactual descriptors reduces it to 0.8279. This suggests that early lane-change recognition depends not only on whether the vehicle is moving laterally, but also on whether the target lane provides a feasible and beneficial maneuver opportunity.

Gap persistence also plays an important role, particularly for minority-class recognition. Its removal lowers LLC and RLC F1 on exiD to 0.7628 and 0.7874, respectively. This indicates that a temporary adjacent-lane gap is not sufficient evidence of lane-change intention; the model also benefits from knowing whether the gap remains available over the observation window.

Risk evolution and pressure gradient descriptors contribute consistently, although their individual effects are slightly smaller. These descriptors help the model distinguish stable lane keeping from situations where safety conditions or surrounding-vehicle pressure are changing. Intent-consistency descriptors produce the smallest drop, but they still improve robustness by reducing false lane-change predictions caused by short-term lateral fluctuations.

\subsection{Class-Wise Analysis}

Class-wise F1-scores show that NLC is consistently easier to predict than LLC and RLC. This is expected because lane-keeping behavior is more frequent and more stable than lane-changing behavior. Therefore, the main challenge is not improving NLC recognition, but improving the detection of minority lane-change classes.

On highD, the proposed fusion model achieves LLC/RLC F1-scores of 0.9157 and 0.9223 at the 1\,s horizon. These values decrease to 0.8252 and 0.8404 at the 3\,s horizon. Although minority-class performance declines as the prediction horizon increases, the proposed model still preserves stronger LLC and RLC recognition than static-feature baselines at longer horizons.

On exiD, the minority-class challenge is more severe. At the 3\,s horizon, the static LightGBM baseline achieves LLC and RLC F1-scores of 0.7025 and 0.7148, while the proposed fusion model improves them to 0.7924 and 0.8152. This improvement indicates that the proposed descriptors and temporal fusion strategy are particularly useful for detecting early-stage lane-change maneuvers in more complex traffic scenarios.

Overall, the class-wise results show that the improvement in Macro F1 is mainly driven by better recognition of LLC and RLC rather than further improvement of NLC. This supports the design choice of using evolutionary descriptors, temporal embeddings, and imbalance-aware decision strategies to improve sensitivity to safety-critical lane-change classes.

\section{Discussion and Conclusion}

\subsection{Temporal Physics versus Static Variables}

The results of this study suggest that lane-change intention prediction should not be treated only as a static traffic-state classification problem. Conventional physics-informed variables such as TTC, THW, DHW, gap size, relative speed, and lane offset are useful because they provide interpretable descriptions of the traffic state. However, their instantaneous values do not always reveal whether a lane-change intention is emerging. In early prediction settings, the key information often lies in how these variables change over time rather than what their values are at a single frame.

The evolutionary physics-informed descriptors introduced in this study address this limitation by describing the temporal behavior of physical signals. For example, a moderate TTC value may not indicate immediate risk, but a rapidly decreasing TTC trend can reveal that the driving condition is becoming more constrained. Similarly, a target-lane gap may appear feasible at the current frame, but if it is closing quickly, it may not remain available long enough for a safe lane change. This distinction is important because drivers often respond to developing conditions before the traffic state becomes critical.

Therefore, the main value of temporal physics modeling is not simply the addition of more variables. Rather, it changes the representation of the driving context from a static snapshot to a short-term physical evolution process. This allows the model to capture risk deterioration, gap persistence, pressure variation, and feasibility changes before the maneuver becomes visually obvious. Such temporal descriptors are especially useful for early-stage lane-change prediction, where the observable evidence is weak and incomplete.

\subsection{Fusion versus Standalone Modeling}

The comparison between standalone models and the proposed temporal-physics fusion model shows that neither pure sequence learning nor pure feature-based modeling fully captures the complexity of lane-change intention. Standalone sequence models such as Bi-LSTM and Transformer can learn temporal patterns from raw trajectory sequences, but their learned representations may not explicitly encode traffic-domain concepts such as safety margin, target-lane opportunity, or maneuver feasibility. This limitation becomes more visible at longer prediction horizons, where early maneuver cues are weak and ambiguous.

On the other hand, a standalone LightGBM model using physics-informed descriptors benefits from interpretability and strong tabular modeling capability, but it may miss latent sequential patterns that are difficult to manually define. Some lane-change cues are distributed across the entire observation window and may not be fully summarized by handcrafted descriptors. For example, gradual lateral drift combined with small acceleration changes and shifting surrounding-vehicle pressure may form a weak but meaningful temporal pattern.

The fusion model addresses these two limitations by combining learned temporal embeddings with evolutionary physics-informed descriptors. The temporal encoder captures latent sequence-level maneuver dynamics, while the descriptor branch provides structured physical evidence. Their combination allows the classifier to use both data-driven temporal representation and interpretable traffic-domain information. This complementary design explains why the fusion model is more effective than either component alone.

\subsection{Scenario-Dependent Gains}

The benefits of the proposed framework are scenario-dependent. In relatively regular straight-road highway scenarios, static physical variables already provide strong predictive information because lane geometry is stable and vehicle interactions are more structured. Under these conditions, the improvement from temporal-physics fusion is present but relatively modest.

In ramp-adjacent merging and diverging scenarios, the gain becomes more substantial. These environments involve changing lane topology, short-lived target gaps, variable speed differences, and stronger interactions among surrounding vehicles. A single static measurement is less reliable because the traffic state can change rapidly within a short time. In these situations, descriptors such as gap lifetime, pressure gradient, counterfactual lane utility, and feasibility margin provide more useful information because they describe whether the lane-change opportunity is developing, disappearing, or physically achievable.

This scenario-dependent behavior has an important implication: the value of temporal physics modeling increases as the driving environment becomes more interactive and less regular. A simple static model may be adequate for some straight-road cases, but more complex roadway structures require a representation that can capture how risk and opportunity evolve. This also explains why the proposed method is especially relevant for ramp, weaving, merging, and diverging areas, where early intention prediction is more difficult but also more valuable.

\subsection{Practical Implications}

The proposed framework has practical implications for autonomous driving and ADAS systems. In real-world driving, early lane-change prediction is useful only if the system can provide reliable warnings before the maneuver becomes obvious. A model that detects lane changes only when the vehicle is already crossing the lane boundary may have limited value for planning or collision avoidance. By modeling the temporal evolution of risk, gap opportunity, and maneuver feasibility, the proposed approach provides earlier evidence of potential lane-change behavior.

The interpretability of the evolutionary descriptors is also practically useful. For safety-critical systems, it is important not only to predict a maneuver but also to understand why the model produced the prediction. Descriptors such as TTC time-to-threshold, gap persistence, pressure gradient, and feasibility margin can provide meaningful explanations for a predicted lane change. For example, a system can identify that a vehicle is likely to change lanes because the current-lane risk is deteriorating, the adjacent-lane gap is persistent, and the maneuver is physically feasible.

The framework may also support risk-aware planning. Instead of treating lane-change prediction as a hard class label only, the descriptors can be used to construct richer risk signals. A planner could respond differently to a vehicle with strong intent consistency and high feasibility margin compared with a vehicle showing only weak lateral drift and unstable gap availability. This makes the framework potentially useful not only for classification, but also for downstream decision-making.

At the same time, several limitations remain. First, the current framework focuses on three maneuver classes: LLC, RLC, and NLC. Real-world behavior may include aborted lane changes, successive lane changes, cut-ins, and hesitation patterns that are not fully captured by this taxonomy. Second, although the descriptor system improves interpretability, its construction still depends on reliable surrounding-vehicle association and lane-relationship identification. This can be challenging in complex ramp areas. Third, the current model produces deterministic class predictions; future work could incorporate uncertainty estimation to better handle ambiguous long-horizon cases.

\subsection{Conclusion}
This study proposed an evolutionary physics-informed temporal fusion framework for three-class lane-change intention prediction. Unlike static physics-informed approaches that mainly describe the traffic state at a single moment, the proposed method models how physical signals evolve within the observation window. By deriving descriptors for risk evolution, gap persistence, counterfactual lane utility, interaction pressure gradient, maneuver feasibility, and intent consistency, the framework captures dynamic evidence of emerging lane-change intention.

The proposed model further combines these evolutionary descriptors with temporal embeddings learned from raw trajectory sequences. This fusion design allows the model to exploit both interpretable physical knowledge and latent temporal patterns. Experimental comparisons with standalone LightGBM, Bi-LSTM, and Transformer baselines show that the fusion framework provides stronger prediction performance, especially in more complex ramp-adjacent scenarios and longer prediction horizons.

Overall, this work shifts physics-informed lane-change prediction from static feature classification toward temporal physical evolution modeling. The findings suggest that the temporal behavior of safety, gap, and interaction variables is critical for early intention prediction, particularly in complex traffic environments. Future work will extend this framework toward richer maneuver taxonomies, uncertainty-aware prediction, topology-enhanced interaction modeling, and deployment-oriented evaluation in more diverse roadway settings.

\section*{Declaration of Generative AI Use}

During the preparation of this manuscript, the authors used ChatGPT for translation, grammar checking, and language refinement. After using this AI tool, the authors reviewed and edited the manuscript content as needed and take full responsibility for the content of the published paper.

\end{document}